# Advancing Multi-Context Systems by Inconsistency Management[*]

Antonius Weinzierl

Institute of Information Systems
Vienna University of Technology
Favoritenstraße 9-11, A-1040 Vienna, Austria
weinzierl@kr.tuwien.ac.at

**Abstract.** Multi-Context Systems are an expressive formalism to model (possibly) non-monotonic information exchange between heterogeneous knowledge bases. Such information exchange, however, often comes with unforseen side-effects leading to violation of constraints, making the system inconsistent, and thus unusable. Although there are many approaches to assess and repair a single inconsistent knowledge base, the heterogeneous nature of Multi-Context Systems poses problems which have not yet been addressed in a satisfying way: How to identify and explain a inconsistency that spreads over multiple knowledge bases with different logical formalisms (e.g., logic programs and ontologies)? What are the causes of inconsistency if inference/information exchange is non-monotonic (e.g., absent information as cause)? How to deal with inconsistency if access to knowledge bases is restricted (e.g., companies exchange information, but do not allow arbitrary modifications to their knowledge bases)? Many traditional approaches solely aim for a consistent system, but automatic removal of inconsistency is not always desireable. Therefore a human operator has to be supported in finding the erroneous parts contributing to the inconsistency. In my thesis those issues will be adressed mainly from a foundational perspective, while our research project also provides algorithms and prototype implementations.

## 1 Introduction

Multi-Context Systems (MCSs) are an expressive formalism for (possibly) non-monotonic knowledge exchange between heterogeneous knowledge sources. These sources are called contexts and formalized as abstract 'logics'. Information flow between contexts is specified using bridge rules which look and behave similar to rules in non-monotonic logic programming (cf. [15]):

$$(k : s) \leftarrow (c_1 : p_1), \ldots, (c_j : p_j), not(c_{j+1} : p_{j+1}), \ldots, not(c_m : p_m). \quad (1)$$

Such a rule states that information $s$ is added to context $k$ if for $1 \leq i \leq j$ knowledge $p_i$ is present in context $c_i$ and for $j + 1 \leq i \leq m$ knowledge $p_i$ is absent in $c_i$. Following common terminology $p_1, \ldots, p_m$ are called beliefs (each of their respective context) and $s$ is the head formula of the bridge rule.

[*] Supported by the Vienna Science and Technology Fund (WWTF), grant ICT08-020.

Consider a hospital where a database with patient records, a medical ontology, and an expert system shall be working together giving decision support on patient medications. The MCS framework is a good choice to realize this. Assume for patient Sue, the database knows that a) her X-Ray result indicates pneumonia, b) a certain blood marker is present, and c) she has no known allergies. The ontology imports information on X-Ray and blood tests using bridge rules

$$(C_{onto} : xray(Sue)) \leftarrow (C_{patients} : labresult(Sue, xray)).$$
$$(C_{onto} : marker(Sue)) \leftarrow (C_{patients} : labresult(Sue, marker)).$$

As the ontology contains the axiom $xray \sqcap marker \sqsubseteq atyp\_pneu$ it concludes that Sue has a atypical pneumonaia, severe kind of pneumonia. Finally, the expert system, a logic program containing rules $give\_weak \vee give\_strong :- pneumonia.$ and $give\_strong :- atyp\_pneumonia.$ suggests one out of two kinds of antibiotics if a patient has pneumonia. But it also respects potential allergies by the constraint $:- give\_strong, not\ allowed\_strong$. As Sue has atypical pneumonia, only the strong antibiotic will help, so the logic program suggests this.

Now assume that Sue is allergic to strong antibiotics, a case that actually happens in the real world. Then the expert system can give no valid suggestion as strong antibiotics have to be given, but at the same time they are forbidden to be applied. This results in the whole system having no 'model' satisfying deductions of all knowledge bases and bridge rules. We call such an MCS inconsistent.[1]

By this example, we identify the following *open problems*:

- the inconsistency above is present due to tuples in the database, terminological assertions in the ontology, logic programming rules in the expert system and, a set of bridge rules establishing the information exchange. In what terms should the inconsistency be described and is there a uniform description irrespective of the specific formalisms used in contexts? Non-monotonicity of bridge rules and contexts is an additional challenge to such a description.
- Given such a description it is very likely that multiple ways exist to restore consistency. Removing some bridge rules would make the above example consistent, but also removal of tuples describing lab results. Similarly, addition of new bridge rules could resolve the inconsistency. If multiple options exist, which is the most preferred to restore consistency? Is it possible to do this in a heterogeneous way, i.e., can the designer of an MCS use a formalism of his own choice to specify his preference? Can such preference be given only for specific parts of an MCS and preference for other parts differently expressed?
- In the above example, the inconsistency can be dealt with locally, e.g., the expert system could switch to use paracoherent semantics and the MCS becomes consistent. For MCSs with cyclic information flow, however, this might be impossible as cyclic information flow can be such that each context returns valid belief sets ("models"), but still for the overall system it does not fit together. How far does local inconsistency management help to resolve inconsistency, e.g., for MCSs with acyclic information flow?
- Besides inconsistency, is the MCS framework so versatile as to use other kinds of rules to connect contexts, e.g., SPARQL queries for information exchange?

---
[1] A complete formalisation of this example is available in [12].

As research on these topics has been started two years ago, the rest of this paper will briefly present results adressing above questions. Regarding research methodologies, we built analogies from existing techniques, e.g., Reiter's diagnosis. For algorithms we resorted to reductions to computational logic and meta-reasoning transformations, e.g., preference is handled in this way. Whenever possible, our invented methods are open so that legacy systems may be integrated to achieve certain tasks, e.g., local inconsistency management.

Contributions summary:

- we developed a uniform representation of inconsistency in terms of bridge rules. This representation leads a) to the notion of inconsistency explanation which separates different sources of inconsistency and points out those bridge rules creating inconsistency and b) to the notion of diagnosis which induce all possible repairs of an inconsistent MCS. Notably, both notions coincide on the overall set of bridge rules which are marked 'faulty.
- on top of those notions, we developed a transformation-based technique to allow meta-reasoning on diagnoses of an inconsistent MCS. This allows system designers to express preferences over diagnoses in a formalism of their own choice. The same techinque also allows to filter out undesired diagnoses.
- for local inconsistency management, a generalization of the MCS formalism was developed allowing to use existing methods of inconsistency management locally for a context. The introduced notion of a context manager allows to employ arbitrary knowledge management techniques locally at a context. It is important that the employed manager can change a knowledge base in a broad range and therefore it can also do other operations like view updates, belief revision, logic program updates, etc.
- for above notions the computational complexity also was analysed.
- to show the versatility of the ideas behind MCS, we also introduced a modified notion of MCS where knowledge exchange is specified using SPARQL queries.

Finally, we also implemented prototypes for evaluating MCSs and computing diagnoses and explanations of inconsistent MCSs.

The remainder of this paper is organized as follows: In Section 2 related work is discussed while Section 3 recapitulates the formal semantics of MCS and our basic notion for inconsistency diagnosis/explanation, it is followed by a short presentation of major achievements in the last two years in Section 4. Finally, Section 5 is an outlook on future work.

## 2 Related Work

With the seminal work of [19] and [16] the notion of context has been introduced to artificial intelligence and logic. In these works, a context is a regarded as a certain point of view in which formal reasoning takes place. The Trento school (cf. [17,22]) formalized and improved this understanding of context. It is notable, however, that those first frameworks consider homogeneous, monotonic logics for representing a context. With [9,21] non-monotonicity was introduced to Multi-Context Systems. Although default negation is added to bridge rules, contexts still are homogeneous or monotonic. Only with [7] the framework has been generalized for non-monotonic bridge rules and heterogeneous contexts. This finally allows to use arbitrary knowledge sources that are connected by (possibly) non-monotonic bridge rules. Our research is based on this notion of MCSs.

To deal with inconsistency, in [5] defeasible rules are introduced as a way of establishing information exchange in MCS. Defeasible rules are similar to bridge rules, but their semantics differs as a defeasible rule does not fire if it would cause an inconsistency by doing so. Several algorithms based on preference orders (or argumentation frameworks [4]) have been proposed. Inconsistency is resolved inherently, but no deeper inconsistency analysis is possible. For our hospital example this would mean that some information simply would not be passed along, e.g., forgetting the illness of *Sue*. Most of the proposed algorithms are based on provenance, which means that context internals have to be exhibited to other contexts. A company making profit by allowing third parties to use its knowledge base, however, will not risk its business by providing such information.

Aside from MCS, other areas deal with knowledge integration and its issues. Peer-to-Peer (p2p) systems [24,10] are similar as knowledge sources interchange pieces of information. Although the notion of a peer is very similar to a context in MCS, the essential feature of p2p systems is that peers may leave and join the system arbitrarily. Therefore research seeks to cope with inconsistency by isolating faulty contexts and simply ignore their information instead of analysing the inconsistency and aiming for a consistent system.

Information integration on the other hand deals extensively with issues like constraint violations that stem from the integration of several databases into a single one (cf. [6] for a survey on data fusion). Its main differences to MCS are that the result of data fusion is one single database which usually uses relational algebra for knowledge representation. MCSs, however, require inconsistency management for multiple, heterogeneous knowledge bases which are not restricted to a relational setting.

For many formalisms, methods of inconsistency handling have been invented, e.g., belief revision or possibilistic reasoning (e.g. [3]) for classical logic, paracoherent semantics for logic programs, etc. These methods can resolve inconsistency locally at a context (cf. Section 4), but they can not guarantee a consistent system. Also, most of those methods are only applicable to a specific formalism instead of a heterogeneous non-monotonic system.

## 3 MCS Preliminaries

Each context of an MCS is seen as a knowledge base built on an underlying logic. To capture different kinds of logics, this notion is general and not defined in the bottom-up style of inductive definitions for syntax and semantics. Instead, its approach is top-down, directly working with sets of well-formed formulas (wffs) and models (called belief sets). The semantics of a logic then only maps each set of wffs to a set of belief sets, i.e., the models of the wffs.

Formally, a logic $L = (\mathbf{KB}_L, \mathbf{BS}_L, \mathbf{ACC}_L)$ consists, of the following components: 1) $\mathbf{KB}_L$ is the set of well-formed knowledge bases of $L$ where each element of $\mathbf{KB}_L$ is a set (of formulas). 2) $\mathbf{BS}_L$ is the set of possible belief sets where we assume that each element of $\mathbf{BS}_L$ is a set (i.e.,a model containing all formulas that are considered true). 3) $\mathbf{ACC}_L : \mathbf{KB}_L \to 2^{\mathbf{BS}_L}$ is a function describing the semantics of $L$ by assigning each knowledge base a set of acceptable belief sets. This concept of a logic captures many monotonic and non-monotonic logics, e.g., classical logic, description logics, modal, default, and autoepistemic logics, circumscription, and logic programs under the answer set semantics.

A *Multi-Context System* $M = (C_1, \ldots, C_n)$ is a collection of contexts $C_i = (L_i, kb_i, br_i)$, $1 \leq i \leq n$, where $L_i = (\mathbf{KB}_i, \mathbf{BS}_i, \mathbf{ACC}_i)$ is a logic, $kb_i \in \mathbf{KB}_i$ a knowledge base, and $br_i$ is a set of bridge rules of form (1) over logics $(L_1, \ldots, L_n)$. Furthermore, for each bridge rule $r \in br_i$ its head formula $s$ is compatible with $C_i$, i.e., for each $H \subseteq \{s \mid r \in br \text{ and } (i : s) \text{ is the head of } r\}$ holds $kb \cup H \in \mathbf{KB}_{L_i}$.

A belief state $S = (S_1, \ldots, S_n)$ of an MCS $M = (C_1, \ldots, C_n)$ is a belief set for every context, i.e., $S_i \in \mathbf{BS}_i$ for all $1 \leq i \leq n$. The semantics of MCS is defined in terms of equilibria, i.e., belief states that reproduce themselves under the application of bridge rules. Formally, let $M$ be an MCS, $C_i$ a context of $M$ and $S = (S_1, \ldots, S_n)$ a belief state of $M$, then an bridge rule $r$ of form (1) is applicable wrt. $S$, denoted by $S \models body(r)$, iff $p_\ell \in S_{c_\ell}$ for $1 \leq \ell \leq j$ and $p_\ell \notin S_{c_\ell}$ for $j < \ell \leq m$. Let $app_i(S) = \{hd(r) \mid r \in br_i \wedge S \models body(r)\}$ denote the heads of all applicable bridge rules of context $C_i$ under $S$, then $S = (S_1, \ldots, S_n)$ is an *equilibrium* of $M$ if and only if $S_i \in \mathbf{ACC}_i(app_i(S))$ for $1 \leq i \leq n$.

**Basic Notions for Inconsistency Analysis (cf. [12]):** We call an MCS $M$ *inconsistent* iff no belief state of $M$ is an equilibrium. To analyse and explain the inconsistency in an MCS, two notions have been developed: consistency-based diagnosis and entailment-based inconsistency explanation. Both notions use bridge rules to characterize 'faulty' information exchange. Intuitively, a diagnosis states how an inconsistent MCS can be changed to get a consistent system and an explanation shows what parts of the system create the inconsistency.

For an MCS $M$, $br_M$ denotes the set of all bridge rules occuring in $M$, $M[R]$ denotes a modified MCS where all bridge rules of $M$ are replaced by those of $R$, and $M \models \bot$ denotes that $M$ is inconsistent. Given an MCS $M$, a *diagnosis* of $M$ is a pair $(D_1, D_2), D_1, D_2 \subseteq br_M$, s.t. $M[br_M \setminus D_1 \cup heads(D_2)] \not\models \bot$. An *explanation* of $M$ is a pair $(E_1, E_2)$ of sets $E_1, E_2 \subseteq br_M$ of bridge rules

s.t. for all $(R_1, R_2)$ where $E_1 \subseteq R_1 \subseteq br_M$ and $R_2 \subseteq br_M \setminus E_2$, it holds that $M[R_1 \cup heads(R_2)] \models \bot$.

For a concise characterization, one usually focuses on subset-minimal diagnoses and explanations. The basic ideas behind both notions appear also in Reiter's seminal work on diagnosis [20]. Our diagnosis is similar to his notion and our explanation is similar to (minimal) inconsistent sets. For differences, we assume the source of inconsistency to be some faulty information exchange, so we only consider bridge rules, and because of the non-monotonic nature of MCSs, a bridge rule can be faulty by firing when it should not and also by not firing when it should. In classical diagnosis, only the former is relevant as monotonic logics only become inconsistent by that. The set of minimal diagnoses can also be seen as describing all minimal repairs, while the set of minimal explanations show hows inconsistency is caused in the system. The set $E_2$ in an explanation also shares some ideas with consistency restoring rules (cf. [2]) of logic programs.

## 4 Contributions: Methods of Inconsistency Management

This section presents contributions and answers the motivational questions raised in the introduction. These are the major published results of my graduate research. Note that authors are listed alphabetically for the respective publications.

**Inconsistency Assessment:** Having jointly developed and investigated, the basic notions for inconsistency analysis, the next step was developing methods to assess inconsistency qualitatively, i.e., filter diagnoses with undesired properties and select most preferred ones. In the spirit of MCS, we do not apply a specific formalism for preference or filters on diagnoses, but rather show how a transformation of the MCS and slight adaption of the notion of diagnosis is sufficient to achieve the desired effects in [13].

As one of the strengths of MCS is the ability to allow arbitrary formalisms for knowledge representation inside contexts, we do not want to restrict the users to a specific kind of representation of filters (or preferences). We therefore devised a meta-reasoning transformation which allows certain contexts to observe which diagnosis is applied to the MCS. The desired filter then is realized inside such an observer context (in a formalisms which is best suited for this task). So an MCS $M$ is transformed into an MCS $M_f$ where an additional observer context $ob$ is added together with some additional bridge rules (details cf. [13]). As $M_f$ contains all contexts and bridge rules of $M$, every diagnosis of $M$ can also be applied to $M_f$. If $ob$ detects an undesired diagnosis $D'$, then $ob$ simply becomes inconsistent, i.e., having no acceptable belief set. Therefore $D'$ is no diagnosis of $M_f$, but all other diagnoses of $M$ are diagnoses of $M_f$. This allows to compute all filtered diagnoses with the same algorithm as for computing subset-minimal diagnoses and it also allows to specify the filter in any desired formalism.

The meta-reasoning transformation also can be applied for multiple observation contexts where each observer only sees some bridge rules instead of all, thus preserving information hiding. As a similar meta-reasoning transformation can be used for comparison of diagnoses, it is possible to realize any given preference order on diagnoses and select the most preferred one. In general, however, this requires exponentially many more bridge rules in the transformed system, but for restricted classes of preference orders it is feasible.

**Inconsistency management at the level of contexts:** For many specific logics and knowledge formalisms, solutions to deal with inconsistency have been developed in the past, e.g., belief revision and paraconsistency for logics, paracoherent logic programming for logic programs, etc. For contexts using the underlying formalism it is desireable that MCSs also offer the same methods of inconsistency handling. Those methods, however, require to modify a knowledge base in more ways, than just the addition of formulas as bridge rules can do.

We therefore propose *managed Multi-Context Systems* (mMCS) in [8] where each context is equiped with a manager that can apply arbitrary changes to the context's knowledge base. Bridge rules in an mMCS are like those of MCS, but their head contains a unary command *op*, e.g., *revise*($s$), *delete*($s$), *add*($s$), to apply the resp. operation on the formula $s$ and the knowledge base of the context.

Managed MCS are a significant generalization of MCS as management functions can be used to realize a multitude of tasks: belief revision, view updates, updates of logic programs. To us, the most interesting is to ensure that contexts have a 'model' for any input. Such contexts are called *totally coherent*. Most notably even mMCS with totally coherent contexts cannot guarantee that the overall system has an equilibrium, but they ensure that inconsistency is only caused by odd-cyclic information flow. It directly follows that any acyclic mMCS with totally coherent contexts is consistent, thus proving local inconsistency management sufficient for acyclic MCS.

**Beyond bridge rules:** In [23] we introduce MCS where knowledge exchange is realised using SPARQL construct-queries. This is surprisingly simple and again shows the versatility of MCS. The resulting SPARQL-MCS framework is related to the MWeb approach [1], but our treatment of variables is different.

## 5 Future Work

As shown above, we were able to answer several foundational questions, give a uniform representation of inconsistency in heterogeneous MCSs, an open integration of preference-based inconsistency assessment, investigating the impact of local inconsistency handling, and making the MCS formalism capable of dealing with arbitrary changes to the knowledge bases of an MCS.

To evaluate the feasibility of the developed methods, we also aim for a reference application which is currently in the making: querying of a DNA database posing questions in (almost) natural language using an ontology and answer-set programs. Intital steps towards exchanging large amounts of information (cf. [14]) also showed that more specialised algorithms are needed.

Investigations whether approximation operators of [11] for logic programs can be translated to MCSs and transferring optimisations for abductive diagnosis (e.g.,[18]) to MCSs are also open tasks.

## 6 Acknowledgements

I am very grateful to my advisor Thomas Eiter, the principal investigator of our research project Michael Fink, and my colleague Peter Schüller who provided guidance, and helped with many fruitful discussions. Thank you.